\documentclass{article} 
\usepackage{ILS-SUMM,times}
\usepackage{graphicx} 
\usepackage{wrapfig,lipsum}

\usepackage{algpseudocode,algorithm,algorithmicx}
\usepackage{mathtools}
\usepackage{amsmath,amsthm,amssymb} 
\usepackage{subcaption}

\usepackage{hyperref}
\usepackage{url}
\usepackage{authblk}

\DeclareMathOperator*{\LocalSearch}{Local-Search-SUMM}
\DeclareMathOperator*{\f}{f}
\DeclareMathOperator*{\N}{N}
\DeclareMathOperator*{\Select}{Select}
\DeclareMathOperator*{\Perturbation}{Perturbation}
\DeclareMathOperator*{\AcceptanceCriterion}{Acceptance-Criterion}
\DeclareMathOperator*{\MMax}{M\_MAX}
\DeclareMathOperator*{\SBest}{S\_Best}
\DeclareMathOperator*{\maxtrials}{MAX\_TRIALS}

\title{ILS-SUMM: Iterated Local Search for Unsupervised Video Summarization} 

\author[1]{\textbf{Yair Shemer}}
\author[2]{\textbf{Daniel Rotman}}
\author[1]{\textbf{Nahum Shimkin}}
\affil[1]{Faculty of Electrical Engineering, Technion, Haifa, Israel}
\affil[2]{IBM Research, Haifa, Israel}
\affil[ ]{sy@campus.technion.ac.il, danieln@il.ibm.com, shimkin@ee.technion.ac.il}

\arxivfinalcopy 
\begin{document}

\maketitle

\begin{abstract}
In recent years, there has been an increasing interest in building video summarization tools, where the goal is to automatically create a short summary of an input video that properly represents the original content. We consider shot-based video summarization where the summary consists of a subset of the video shots which can be of various lengths. A straightforward approach to maximize the representativeness of a subset of shots is by minimizing the total distance between shots and their nearest selected shots. We formulate the task of video summarization as an optimization problem with a knapsack-like constraint on the total summary duration. Previous studies have proposed greedy algorithms to solve this problem approximately, but no experiments were presented to measure the ability of these methods to obtain solutions with low total distance. Indeed, our experiments on video summarization datasets show that the success of current methods in obtaining results with low total distance still has much room for improvement. In this paper, we develop ILS-SUMM, a novel video summarization algorithm to solve the subset selection problem under the knapsack constraint.  Our algorithm is based on the well-known metaheuristic optimization framework -- Iterated Local Search (ILS), known for its ability to avoid weak local minima and obtain a good near-global minimum. Extensive experiments show that our method finds solutions with significantly better total distance than previous methods. Moreover, to indicate the high scalability of ILS-SUMM, we introduce a new dataset consisting of videos of various lengths.

\end{abstract}

\section{Introduction}
In recent years, the amount of video data has significantly increased. In addition to many cinema movies, news videos, and TV-shows, people frequently shoot events with their cellphone and share it with others on social media.
To illustrate, it has been reported that every minute, 300 hours of new videos are uploaded to YouTube.
Consequently, the user's ability to manage, search, and retrieve a specific item of content is limited.
One remedy to this challenge can be an automatic video summarization algorithm where the goal is to create a shortened video that contains the essence of the original video. If fact, several commercial video summarization products are already on the market. 

\par
Various video summarization algorithms have been suggested in the literature.
In most methods, the process consists of two main stages -- segmenting the video into short video shots, and then choosing a subset of the shots to aggregate a summary \citep{otani2019rethinking}.
In order to be a good summary, this shot subset selection should optimize a certain property.
For example, the selected shots should well represent the content of the video in the sense that each object from the original video has a similar object in the summary.
\par
Video summarization approaches can generally be divided into supervised and unsupervised.
Supervised methods include exploiting a ground truth importance score of each frame to train a model \citep{zhang2016video,zhao2018hsa,gygli2015video} and utilizing auxiliary data such as web images \citep{khosla2013large}, titles \citep{song2015tvsum}, category \citep{potapov2014category} and any other side information \citep{yuan2017video}.
A pitfall of a supervised approach is the necessity of expensive human-made labels.
This drawback is especially restrictive because of the complicated and vague structure of a good summary, which requires a lot of labeled data.
\par
In contrast, unsupervised methods do not need human-made labels as they follow rational guidelines for creating a good summary.
One group of unsupervised algorithms maximizes the similarity between the original video and the generated summary using generative adversarial networks as evaluators \citep{mahasseni2017unsupervised,jung2019discriminative,yuan2019cycle}, or by dictionary learning \citep{cong2011towards,zhao2014quasi}.
Another salient group of methods seek to minimize the total distance between shots and their nearest selected shots while satisfying the limit on the summary duration.
Attempts in this direction include using submodular optimization \citep{gygli2015video}, reinforcement learning \citep{zhou2018deep} and clustering methods \citep{chheng2007video,de2011vsumm,hadi2006unsupervised}.
\par
Even though several methods have been proposed for minimizing this total distance, no experiments have presented to directly measure the success of these methods in obtaining solutions with low total distance.
Our experiments indicate that the best current existing method obtains solutions with total distance, which is, in some datasets around 10\% worse than the optimal solution on average.
Hence, we see that there is room for a new method that leads to better solutions.
\par
In this paper, we propose ILS-SUMM, a novel unsupervised video summarization algorithm which uses the Iterated Local Search (ILS) optimization framework \citep{lourencco2003iterated} to find a representative subset of shots.
We formalize the following optimization problem: given the entire set of shots with varied shot duration, select a subset which minimizes the total distance between shots and their nearest selected shots while satisfying a knapsack constraint, i.e., the limit on the summary duration. This problem is known in the Operations Research field as the Knapsack Median  (KM) problem, and is known to be NP-hard. A major challenge in performing a local search in the solution domain is the high chance of getting stuck in a local minimum because of the hard knapsack constraint \citep{fang2002reducing}. Therefore we use the ILS framework, which is the basis for several state-of-the-art algorithms for NP-hard problems \citep{lourencco2019iterated}.
\par
ILS-SUMM creates a video summary by selecting shots that well represent the original video, using the ILS framework. First, it initializes a solution that satisfies the summary duration limit. Then, it applies steps of improvements by adding or replacing one shot at a time, while allowing only steps that obey the knapsack constraint. When a local minimum point is reached, the ILS executes a gentle, yet noticeable, perturbation to the current solution, to get out from the local minimum while trying to keep part of the high quality of the solution it obtained. We perform extensive experiments on the SumMe and TvSum benchmarks showing that our method finds solutions that are on average less than 2\% worse than the optimal solution, which is significantly superior than the results of previous methods. Moreover, experiments on long real open-source movies indicate ILS-SUMM scalability. A Python implementation of the proposed method is released in [https://github.com/YairShemer/ILS-SUMM].

\section{Related Work}
\subsection{Video Summarization}
Various unsupervised video summarization methods have been presented in the recent literature.
Most methods share the underlying assumption that a good summary should represent and be similar to the original video. \citet{cong2011towards} and \citet{zhao2014quasi} build a representative dictionary of key frames that minimizes the reconstruction error of the original video. \citet{mahasseni2017unsupervised} train a deep neural network to minimize the distance between original videos and a distribution of their summarizations.
\par
\citet{chheng2007video} cluster the video shots into k clusters using the k means algorithm, and then select shots from various clusters. \citet{gygli2015video} apply submodularity optimization to minimize the total distance between shots and the nearest of the selected shots. \citet{zhou2018deep} use a reinforcement learning framework to train a neural network to select frames such that the representativeness and diversity of the summary will be maximized.
\par 
In recent years, most methods evaluate their results by measuring the similarity between automatic summaries and human-made summaries. Recently, \citet{otani2019rethinking} observed that randomly generated summaries obtain competitive or better performance in this metric to the state-of-the-art. Based on this surprising observation, instead of evaluating our method using human labels, we measure the success of our algorithm in terms of having low total distance between all shots and the nearest of the selected shots.

\subsection{Iterated Local Search}
In the ILS framework, a sequence of locally optimal solutions is iteratively generated by a heuristic algorithm. The initial point of the heuristic search in each iteration is a perturbation of a previous obtained solution, rather than a complete random trail. 
ILS has been applied successfully to various optimization problems, leading to high performance and even established the current state-of-the-art algorithms in some tasks. 
\par
Some successful applications of ILS include solving common combinatorial problems such as graph partitioning problems, traveling salesperson problems and scheduling problems \citep{de2016iterated}, in addition to other applications such as image registration \citep{cordon2006image}, car sequencing \citep{cordeau2008iterated,ribeiro2008hybrid} and the generalized quadratic multiple knapsack problem \citep{avci2017multi}.
To the best of our knowledge, this paper is the first to apply Iterated Local Search to the Knapsack Median problem.

\section{Method}
In this section, we introduce an unsupervised method for video summarization, based on the Iterated Local Search (ILS) optimization framework.
\subsection{Formulation}
Given an input video ${v}$, the video is divided temporally into a set of shots ${S_{v}} = \left\{ {{s_1},{s_2}, \ldots {s_N}} \right\}$, where $N$ is the number of shots in the video. We denote the duration in seconds of a shot $s$ as $t\left( s \right)$. Each shot is represented by its middle frame feature vector $x\left( s \right)$ (for details see section 4.1). A condensed video summary is a representative subset of shots $\;{S_{summ}} \subseteq {S_{v}}$. The summarization task is then formulated by the following optimization problem.
We denote $TD(S_{summ}|S_v,x(s))$ as the total distance between all video shots and their nearest shot in $S_{summ}$:
\begin{align} \label{eq:TD objective function}
    TD(S_{summ}|S_v,x(s))=\mathop \sum \limits_{s' \in {S_{v}}} \mathop {\min }\limits_{s \in S_{summ}} \{ {{\rm{dist}}\left( {x\left( {s'} \right),x\left( s \right)} \right)} \}
\end{align}
The objective is to obtain the subset $S_{summ}$ that minimizes the total distance:
\begin{align} \label{eq:KM objective function}
    S_{summ}^* = \mathop {{\rm{argmin}}}\limits_{{S_{summ}} \subseteq {S_{v}}}TD(S_{summ}|S_v,x(s)),
    \end{align}
subject to:
\begin{align} \label{eq:KM constraint}
    \mathop \sum \limits_{s \in {S_{summ}}} t\left( s \right) \le T ,
\end{align}
where $\rm{dist}\left(x,y\right)$ denotes some distance metric between x and y, and T is the maximum total duration in seconds allowed for the video summary. Equation \ref{eq:KM objective function} expresses the goal that the subset will minimize the total distance between shots and their nearest selected shots. Equation \ref{eq:KM constraint} encodes the knapsack constraint which limits the total duration of the selected shots not to exceed T seconds. This problem is known in the Operations Research field as the Knapsack Median (KM) problem. 
\par
Motivated by the success of Iterated Local Search (ILS) in many computationally hard problems \citep{lourencco2019iterated}, we use this simple yet powerful metaheuristic to address the KM problem, and consequently obtain a representative video summary. 
A fundamental component in the ILS framework is a local search algorithm. In the following section, we first introduce a local search algorithm tailored to the KM problem, which we name Local-Search-SUMM, and subsequently we present the complete ILS-SUMM algorithm.
\subsection{Local-Search-SUMM}
A local search algorithm starts from a particular solution of a given problem and sequentially improves the solution by performing a local move at each step \citep{osman1997meta}. The pseudo-code of Local-Search-SUM, a local search algorithm we developed for the KM problem is given by Algorithm \ref{alg:Local-Search-SUMM}. This algorithm contains the following functions: $\f()$ -- an objective function, $\N()$ -- a map between a solution $S$ to a neighborhood of solutions, and $\Select()$ -- a function that selects one of the neighbors, all detailed below. As an input, Local-Search-SUM gets $S_{init}$ -- an initialization of a solution. In each iteration of the algorithm, it selects a neighbor of the current solution and moves to this neighbor if it decreases the objective function. The loop continues until a local minimum is reached or until $\maxtrials$, i.e., the predefined maximum number of trials, is reached. To solve the KM problem with a local search algorithm, we use the setting as described below.

\begin{algorithm}[t]
\caption{Local-Search-SUM}\label{alg:Local-Search-SUMM}
\begin{algorithmic}[1]
\State $\textbf{Input:} \ S_{init}, S_v, x(s), t(s), T$
\State $S^*\coloneqq S_{init}$
\For{$k \coloneqq 1\text{ to }\maxtrials$}
    \State $S\coloneqq \Select(\N(S^*, S_v, t(s) ,T), x(s))$
    \If{$\f(S|S_v, x(s))<\f(S^*|S_v, x(s))$}
        \State $S^*\coloneqq S$
    \EndIf
\EndFor
\State{\textbf{return: $S^*$}} 
\end{algorithmic}
\end{algorithm}

\begin{algorithm}[t]
\caption{ILS-SUMM}\label{alg:ILS-SUMM}
\begin{algorithmic}[1]
\State $\textbf{Input:} \ S_{init}, S_v, x(s), t(s), T$
\State $S^*\coloneqq \LocalSearch(S_{init}, S_v, x(s), t(s), T)$
\State $\SBest \coloneqq S^*$
\State $M \coloneqq 1$
\While{$M \leq \MMax$}
    \State $S'\coloneqq \Perturbation(S^*,S_v,M,t(s),T)$
    \State $S'^*\coloneqq \LocalSearch(S', S_v, x(s), t(s), T)$
    \State $S^*\coloneqq \AcceptanceCriterion(\SBest, S'^*, S^*, S_v, x(s))$
    \If{$\f(S'^*|S_v, x(s)) < \f(\SBest|S_v, x(s))$}
        \State $\SBest \coloneqq S'^* $
        \State $M \coloneqq 1 $
    \Else
        \State $M \coloneqq M+1$
    \EndIf
\EndWhile
\State{\textbf{return: $\SBest$}}
\end{algorithmic}
\end{algorithm}

\par
\textbf{Objective function.} Straightforwardly, we define the objective function as the total distance between shots and their nearest selected shots:
\begin{align}
    f(S_{summ}|S_v,x(s))=TD(S_{summ}|S_v,x(s))
\end{align}
Note that an extension to a multi-objective function is straightforward, by changing $f()$ to be a weighted sum of objectives as proposed by \citet{gygli2015video}. 
\par
\textbf{Initialization.} To deal with the knapsack constraint, we define the local search initialization and neighborhood such that throughout all the solution process, only feasible solutions will be considered. To ensure that the initialized solution $S_{init}$ satisfies the knapsack constraint, the initialization subset is set to be the single shortest shot.
\par
\textbf{Neighborhood.} A neighborhood $\N(S_{summ}, S_v, t(s),T)$ includes any subset which is obtained by swapping or adding a single shot to $S_{summ}$, while satisfying the knapsack constraint. Removing a shot will never decrease the objective function, and therefore is not included in the neighborhood set.  
\par
\textbf{Selection.} As a selection method, $\Select\left(\N(), x(s) \right)$, we use the steepest descent method, i.e., selecting the neighbor which decreases the objective function the most. To boost run-time performance the algorithm first considers adding a shot, and only if this is impossible it considers swaps. This approach reduces the complexity of the algorithm and demonstrates significantly better run-time performance in our experiments.

\subsection{ILS-SUMM}
\begin{wrapfigure}{r}{7cm}
    \centering
    \includegraphics[width=7cm]{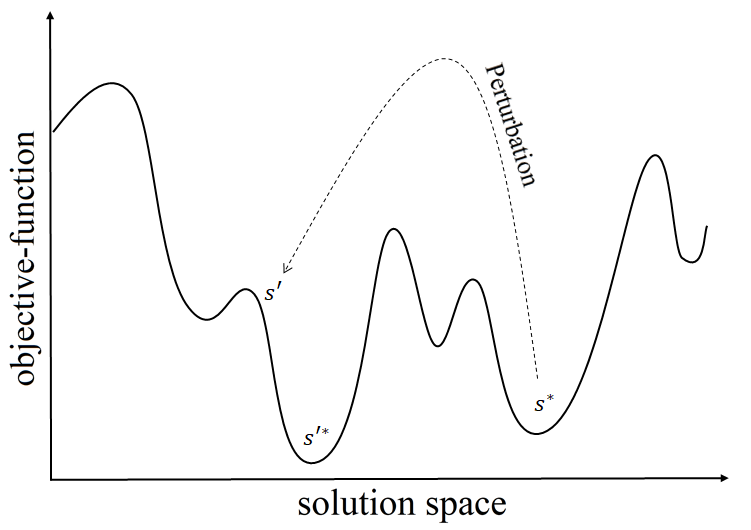} 
    \caption{Illustration of Iterated Local Search framework. Given a local minimum $S^*$, a perturbation leads to a solution $S'$. Then, a call of a local search algorithm obtains a local minimum $S'^*$ which is potentially better than $S^*$. }
    \label{fig:Illustration of ILS}
\end{wrapfigure}

A local search algorithm may lead to a poor local minimum that is far away from the global minimum. Hence, after getting stuck in a local minimum, it is worthwhile to continue searching for other solutions which potentially can be far better. ILS performs this continued search, by repeatedly calling a local search algorithm which in each call starts from a different starting solution. As illustrated in figure \ref{fig:Illustration of ILS}, in each iteration the starting point of the next iteration is a perturbation of the current solution.

\begin{algorithm}[t]
\caption{Perturbation}\label{alg:PERTURBATION}
\begin{algorithmic}[1]
    \State $\textbf{Input:\ } S, S_v, M, t(s), T$
    \State $S' \xleftarrow{} S$
    \For{$m \coloneqq 1\ to\ M$}
        \State $S' \xleftarrow{} S' \cup \mathop{{\arg\min}} \limits_{s \in S_v|s \notin S'} t(s) 
        \backslash \mathop{{\arg\max}} \limits_{s \in S'} t(s) $
    \EndFor
    \If{$\mathop \sum \limits_{s \in {S'}} t\left(s \right) > T$}
    \State $S' \xleftarrow{} S$
    \EndIf
    \State{return: $S'$} 
    \end{algorithmic}

\end{algorithm}

ILS-SUMM pseudo-code is given by  Algorithm \ref{alg:ILS-SUMM}. It consists of three main components which are executed at each iteration: A perturbation of the last solution, a call of a local search algorithm and a decision whether to accept the new found local minimum or to stay with the old solution. As a local search, we use the Local-Search-SUMM introduced above. In the following, we will go into more details regarding the perturbation mechanism and acceptance criterion.

\par
\textbf{Perturbation.} In this stage, the previous solution is modified to a different solution. Specifically, ILS-SUMM perturbs a given subset $S$ by swapping $M$ shots in $S$ with $M$ shots that are currently not in $S$. See the perturbation mechanism pseudo-code in Algorithm \ref{alg:PERTURBATION}. To maximize the chance of getting a feasible solution, the perturbation is executed in a constraint-greedy manner. This constraint greediness means that the longest-duration currently selected shots are swapped with the shortest-duration non-selected shots. However, if the new solution does not satisfy the knapsack constraint, then the original solution is returned (line 7). 
This perturbation mechanism is deterministic. Another option is to add randomness when selecting which shots to swap. Since the stochastic version did not lead to an improvement in the experiments, we retain the deterministic version which also enjoys the benefit of repeatability. 

\par
The strength of the perturbation can range between two extremes. On one extreme, it can totally change the solution and in fact, restart from a new solution. These complete initializations typically lead to long iterations and poor solutions because of the low quality of the starting solution.
On the other extreme, applying a weak perturbation which only slightly changes the solution may lead to being stuck repeatedly in the same local minimum; hence, a good perturbation has a balanced intensity. As described in the ILS-SUMM pseudo code (Algorithm \ref{alg:ILS-SUMM}), we use a gradually increasing perturbation strength. It starts from $M=1$ and gradually increases $M$ by one until $\MMax$, i.e., a predefined maximum value of $M$ is reached. In this way, we use the minimal strength of perturbation that accomplishes exiting the current local minimum.   
\par
\textbf{Acceptance Criterion.} In this stage, the algorithm decides which solution will be perturbed in the next iteration to get a new starting point. Two extreme cases of this procedure are either to always continue with the new local minimum obtained or to stick with the currently best achieved local minimum. The first extreme can be interpreted as a random walk over the local minima, whereas the second extreme can be viewed as a greedy local search over the local minima  \citep{lourencco2019iterated}. An intermediate approach would be to prioritize good solutions while occasionally exploring inferior solutions. An example of such a scheme is the Metropolis Heuristic where worse solutions randomly get the chance to be explored. An interesting modification of the Metropolis Heuristic is Simulated Annealing \citep{van1987simulated}, where the temperature of these exploration events, i.e., the probability of moving to a worse solution, progressively decreases. Since all the above options demonstrate similar results in our experiments, we assign an acceptance criterion which chooses the best achieved local minimum.

\section{Experiments}
\subsection{Experimental Setup}

\textbf{Datasets.} We evaluate our approach on two popular video summarization datasets -- SumMe \citep{gygli2014creating} and TvSum \citep{song2015tvsum}, as well as on our new Open Source Total Distance dataset with full-length videos which we present below.
SumMe consists of 25 user videos on various topics such as cooking, traveling, and sport. Each video length ranges from 1 to 6 minutes. TvSum consists of 50 videos with a duration varying between 2 to 10 minutes.
We use a video shot segmentation technology to subdivide each video into shots (for details see Section 4.1). For SumMe and TvSum we use their common summary length limit which is 15\% of the video length.
We use Python PuLP library to obtain the optimal total distance of each video using an integer programming solver. As we will show in the results section, this integer programming tool has poor run-time scalability, but is useful for obtaining the optimal total distance as ground truth.

To evaluate the total distance results and the scalability on longer movies, we establish a new total distance dataset -- Open-Source-Total-Distance (OSTD). This dataset consists of 18 movies with duration range between 10 and 104 minutes leveraged from the OVSD dataset \citep{rotman2016robust}. For these videos, we set the summary length limit to be the minimum between 4 minutes and 10\% of the video length.
Links for OSTD movies and ground truth optimal total distance of all above datasets are available on [https://github.com/YairShemer/ILS-SUMM]. 

\par
\textbf{Implementation Details.}
For temporal segmentation, we use two different types of shot segmentation methods, in accordance with video types. For the OSTD movies, we use FFprobe Python tool \citep{FFprob} since this tool has high accuracy when applied on videos with fast shots transitions. For the SumMe and TvSum datasets, we use KTS proposed by  \citet{potapov2014category}, since this shot segmentation method is more appropriate for catching slow shot transitions which are common in these two datasets.
For feature extraction, we use the RGB color histogram with 32 bins for each of the three color channels. See some comments on using deep features in section 4.3.
\par
For the perturbation mechanism of the ILS-SUMM, we set the maximum value of $M$ to be 5 since we found that this value leads to a balanced perturbation intensity. However, we observed that ILS-SUMM is not sensitive to this value, and other values are just as satisfactory.
\par
\textbf{Evaluation.} To compare between different approaches, we calculate the total distance defined in equation (1) that each approach achieved for each video. We then calculate the optimality percentage, i.e., the ratio between the optimal value and the achieved value, mutiplied by $100$. For each method, we average all optimality percentages achieved on all the videos of a specific dataset, and report the averaged optimality ratio.
\par
\textbf{Comparison.} To compare total distance results with other approaches, we apply DR-DSN \citep{zhou2018deep} and Submodular \citep{gygli2015video} on the datasets. For both methods, we use the implementations provided by the original authors. Although both methods can optimize multiple objectives, for our experiments we set them to maximize only representativeness since this is the evaluation metric we use. As mentioned above, an extension of our method to a multi-objective setting is straightforward, but to simplify the comparison we focus on representativeness. 
\begin{figure*}[t]
    \centering
    \includegraphics[width=0.99\textwidth]{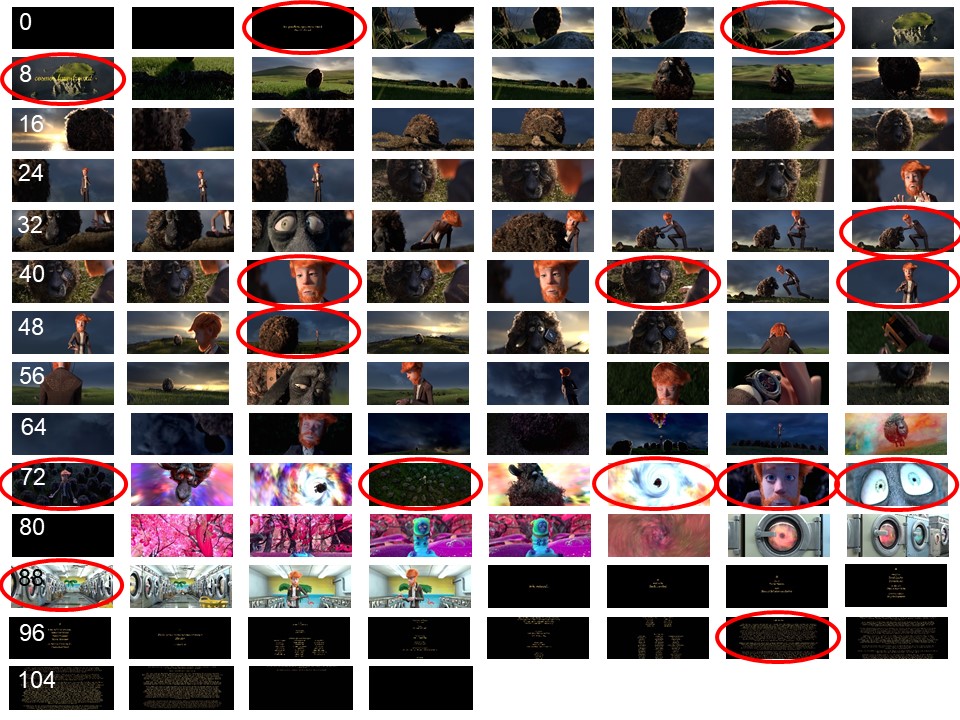} 
    \caption{ILS-SUMM selection in Cosmos Laundromat movie from OSTD data set. The middle frame of shot is presented. A red circle denotes the shots that were chosen by ILS-SUMM algorithm.}
    \label{fig:cosmos}
\end{figure*} %

\subsection{Results}
First, we compare our method with simple local search baselines. Then, we compare our method with previously proposed algorithms.

\textbf{Comparison with baselines.} We set the baseline algorithms as two variants of local search algorithms. The first baseline is Local-Search-SUMM described above in Algorithm \ref{alg:Local-Search-SUMM}. The second baseline, denoted by Restarts-SUMM, repeatedly restarts Local-Search-SUMM initialized with a different single shot at each restart and then selects the best result. The algorithm stops when it finishes going over the entire video shots or when the run-time resources are reached. For each video we set the Restart-SUMM maximum run-time allowed to be the video duration. Table \ref{table:total-SUMM baselines} reports the total distance achieved by the baselines and ILS-SUMM on SumMe, TvSum, and OSTD datasets.

\begin{table}[h]
\begin{center}
    \begin{tabular}{ | c | c | c | c | c | c | }
     \hline
    & \text{SumMe} & \text{TVSum} & \text{OSTD}  \  \\ \hline
    \text{Local-Search-SUMM} & \text{70.80\%} &\text{87.11\%} & \text{91.66\%} \\ \hline
    \text{Restart-SUMM} & \text{93.19\%} & \text{98.19\%} & \text{94.95\%} \\ \hline
    \textbf{ILS-SUMM} & \textbf{98.48\%} & \textbf{99.27\%} & \textbf{98.38\%} \\ \hline
    \end{tabular}
\end{center}
\caption{Results (total distance optimality percentage) of different variants of local search on SumMe, TVSum and OSTD.}\label{table:total-SUMM baselines}
\end{table}

We can see that ILS-SUMM clearly outperforms Local-Search-SUMM. This result demonstrates the importance of the exploration process of ILS, since stopping the algorithm in the first reached local minimum as done in Local-Search-SUMM is far from optimal. 

\par
Although Restart-SUMM is significantly better than Local-Search-SUMM, it is still inferior to ILS-SUMM. More essentially, Restart-SUMM is highly unpractical since in many videos the time it takes for generating a summary with Restart-SUMM is equal to the time it takes watching the full video (for more details see the run-time analysis below). 
This indicates the usefulness of the ILS perturbation mechanism, which rather than initializing the solution to a completely new solution, partially reuses the good solution it already obtained and thus obtains better results in less time.

\par
\textbf{Comparison with previous approaches.} 
Table \ref{table:total-SUMM previose approaches} shows the results of ILS-SUMM measured against other video summarization methods that aim to minimize total distance, on SumMe, TvSum, and OSTD. It can be seen that ILS-SUMM significantly outperforms the other approaches on all datasets.

\begin{table}[h]
\centering
\caption{Results (total distance optimality percentage) of different approaches on SumMe, TVSum and OSTD. Our ILS-SUMM exhibits a significant advantage over others}\label{table:total-SUMM previose approaches}
\begin{tabular}{ | c | c | c | c | c | c | }
\hline
    & \text{SumMe} & \text{TVSum} & \text{OSTD}  \  \\ \hline
    \text{DR-DSN} & \text{90.78\%} & \text{82.50\%} & \text{62.56\%} \\ \hline
    \text{Submodular} & \text{85.18\%} & \text{94.14\% } & \text{95.99\%} \\ \hline
    \textbf{ILS-SUMM} & \textbf{98.48\%} & \textbf{99.27\%} & \textbf{98.38\%} \\ \hline
    \end{tabular}
\end{table}

\par
\textbf{Run-time performance.} 
Table \ref{table:run_time_ILS_SUMM_PuLP} presents the run-time measurements of the PuLP, Submodular, Restart-SUMM and ILS-SUMM methods in OSTD dataset. 
Our experiments demonstrate that for obtaining a reasonable solution, Submodular is the fastest approach.
These results may be expected since Submodular runs only two iterations of greedily adding shots, without any further exploration. However, as we presented above, ILS-SUMM obtains significantly better results than submodular optimization, while enjoying a substantially better run-time scalability than PuLP. With these numbers it is possible to make a decision of solution optimality vs. run-time for a given specific use of video summarization. 

\begin{table}[h]
\centering
\caption{Run-time comparison (\% of video duration) between PuLP, Submodular, Restart-SUMM and ILS-SUMM in OSTD dataset.}\label{table:run_time_ILS_SUMM_PuLP}
\begin{tabular}{ | l | c | c | c | c | }
\hline
	 & \text{PuLP} & \text{Submodular} & \text{Restart-SUMM} & \textbf{ILS-SUMM} \\ \hline
	\text{Big Buck Bunny (596 [Sec])} & \text{1.96\%} & \text{0.02\%} & \text{8.08\%} & \textbf{0.33\%} \\ \hline
	\text{La Chute d’une Plume (624 [Sec])} & \text{0.48\%} & \text{0.01\%} & \text{2.12\%
} & \textbf{0.09\%} \\ \hline
	\text{Elephants Dream (654 [Sec])} & \text{0.60\%} & \text{0.01\%} & \text{2.34\%
} & \textbf{0.11\%} \\ \hline
	\text{Meridian (719 [Sec])} & \text{1.66\%} & \text{0.01\%} & \text{5.10\%
} & \textbf{0.21\%} \\ \hline
	\text{Cosmos Laundromat (731 [Sec])} & \text{0.95\%} & \text{0.01\%} & \text{2.41\%
} & \textbf{0.06\%} \\ \hline
	\text{Tears of Steel (734 [Sec])} & \text{1.51\%} & \text{0.01\%} & \text{6.90\%
} & \textbf{0.29\%} \\ \hline
	\text{Sintel  (888 [Sec])} & \text{0.83\%} & \text{0.01\%} & \text{5.67\%
} & \textbf{0.32\%} \\ \hline
	\text{Jathia's Wager  (1261 [Sec])} & \text{2.09\%} & \text{0.02\%} & \text{21.20\%
} & \textbf{0.38\%} \\ \hline
	\text{1000 Days (2620 [Sec])} & \text{5.48\%} & \text{0.02\%} & \text{71.48\%
} & \textbf{0.98\%} \\ \hline
	\text{Pentagon (3034 [Sec])} & \text{4.71\%} & \text{0.02\%} & \text{50.10\%
} & \textbf{0.60\%} \\ \hline
	\text{Seven Dead Men (3424 [Sec])} & \text{22.49\%} & \text{0.02\%} & \text{62.47\%
} & \textbf{0.36\%} \\ \hline
	\text{Boy Who Never Slept (4186 [Sec])} & \text{25.47\%} & \text{0.03\%} & \text{100\%} & \textbf{0.84\%} \\ \hline
	\text{Sita Sings the Blues (4891 [Sec])} & \text{58.33\%} & \text{0.06\%} & \text{100\%} & \textbf{4.97\%} \\ \hline
	\text{CH7 (5189 [Sec])} & \text{24.10\%} & \text{0.02\%} & \text{100\%} & \textbf{0.85\%} \\ \hline
	\text{Honey (5210 [Sec])} & \text{45.12\%} & \text{0.03\%} & \text{100\%} & \textbf{1.19\%} \\ \hline
	\text{Valkaama (5586 [Sec])} & \text{51.82\%} & \text{0.04\%} & \text{100\%} & \textbf{1.86\%} \\ \hline
	\text{Star Wreck (6195 [Sec])} & \text{91.96\%} & \text{0.05\%} & \text{100\%} & \textbf{2.38\%} \\ \hline
	\text{Route 66 (6205 [Sec])} & \text{49.06\%} & \text{0.05\%} & \text{100\%} & \textbf{3.13\%} \\ \hline
    \end{tabular}
\end{table}

\subsection{Deep Features}
Recently, deep features are being used for many applications, including video summarization, as they give state-of-the-art results in many applications such as semantic image classification, visual art processing and image restoration. However, since the question of what is a right evaluation of video summarization is still an open question \citep{otani2019rethinking}, there is no solid evidence for an advantage in using deep features rather than color histogram features for this task. 
To decide which features to use, we extracted both types of features for all videos from the SumMe dataset. For color histograms we used 32 bins of each of the RGB channels, and as deep features we used the penultimate layer from the ResNet model \citep{he2016deep} pre-trained on ImageNet \citep{deng2009imagenet}. Then, for each video, we applied a dimensionality reduction on these features using PCA.

\begin{figure}[t]
\begin{subfigure}{0.48\textwidth}
  \centering
  \includegraphics[width=\linewidth]{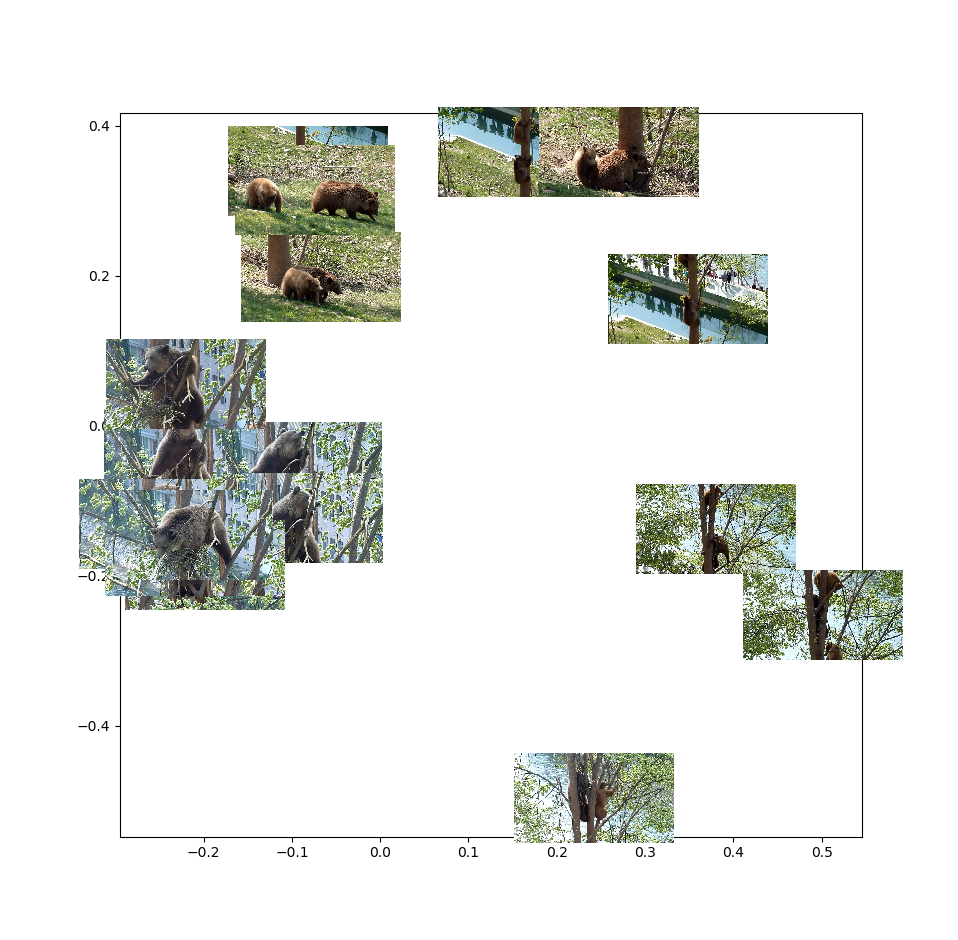}
  \caption{Visualization of the \textbf{color histogram features}.}
  \label{fig:features PCA - shallow}
\end{subfigure}
\begin{subfigure}{0.48\textwidth}
  \centering
  \includegraphics[width=\linewidth]{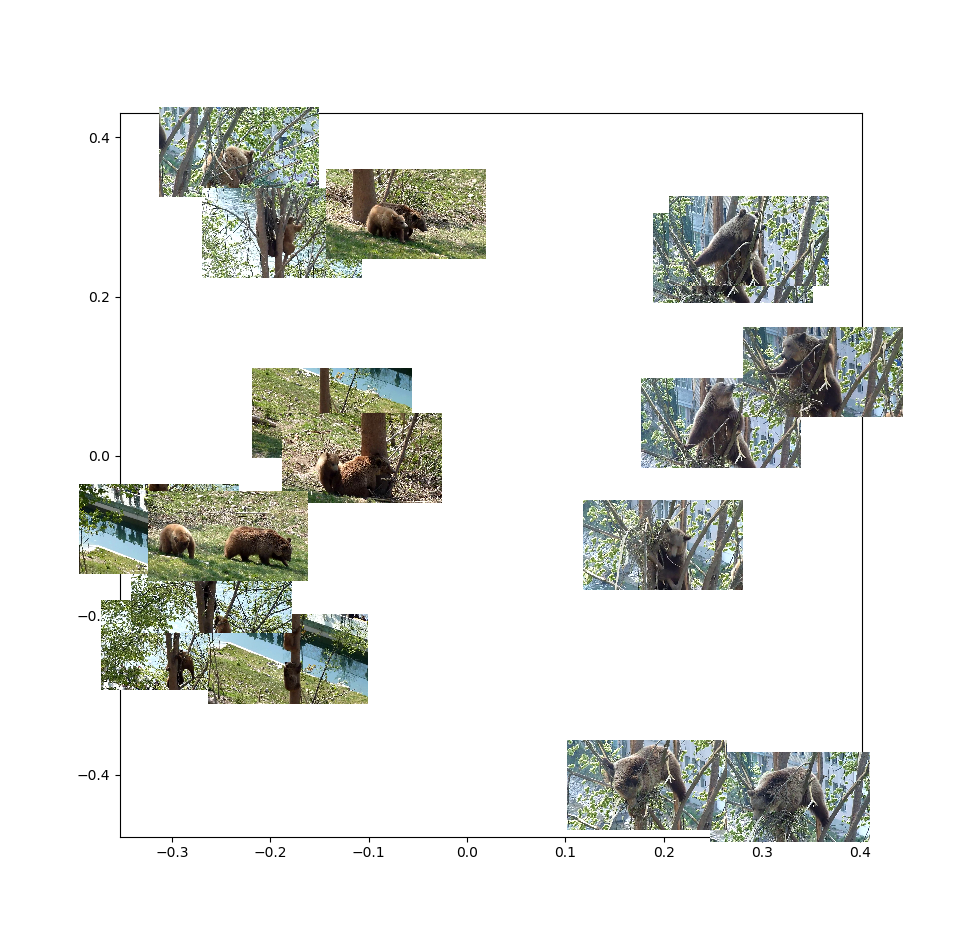}
  \caption{Visualization of the \textbf{deep features}.}
  \label{fig:features PCA - deep}
\end{subfigure}
\caption{Visualization of the features of the shots of "Bearpark Climbing" video from SumMe dataset. In each figure the features dimension were reduced to two dimensions using PCA. Figure (a) visualizes the color histogram features, and Figure (b) visualizes the deep features.}
\label{fig:fig}
\end{figure}

\par
We observed that even though deep features are better in representing the semantics of the images, color histogram features seem to represent  background and scene changes better. For example, Figure \ref{fig:features PCA - shallow} visualizes the color histogram feature space of the "Bearpark Climbing" video from the SumMe dataset, and Figure \ref{fig:features PCA - deep} visualizes the deep feature space. The plot's axes are the two first principle components of the shot features. Each shot is represented by the image of the middle frame in the shot. It can be seen that in both cases, deep and shallow features, different scenes tend to be located and grouped in different parts of the feature space. However, the grouping of the color histogram space visually looks better than the deep features grouping, especially given the task definition of creating a summary which is visually similar to the source. Therefore we used the color histogram features to represents shots in this paper. Future research may examine the integration of deep and color histogram features. 

\section{Conclusion}
In this paper, we have proposed a new subset selection algorithm based on the Iterated Local Search (ILS) framework for unsupervised video summarization. Motivated by the success of ILS in many computationally hard problems, we leverage this method for explicitly minimizing the total distance between video shots and their nearest selected shots under a knapsack-like constraint on the total summary duration.
We have shown that a proper balance between local search and global exploration indeed leads to an efficient and effective algorithm for the Knapsack Median problem.
Our experiments on video summarization datasets indicate that ILS-SUMM outperforms other video summarization approaches and finds solutions with significantly better total distance. Furthermore, experiments on a long videos dataset we have introduced demonstrate the high saclability of our method.

\bibliography{ILS-SUMM}
\bibliographystyle{ILS-SUMM}

\end{document}